\documentclass[runningheads]{llncs}

\usepackage{eccv}

\usepackage{eccvabbrv}

\usepackage{graphicx}
\usepackage{booktabs}
\usepackage{colortbl}
\usepackage{array}
\usepackage{multirow}

\usepackage[accsupp]{axessibility}  

\usepackage{hyperref}

\usepackage{orcidlink}

\begin{document}

\title{HEAD: A Bandwidth-Efficient Cooperative Perception Approach for Heterogeneous Connected and Autonomous Vehicles} 

\titlerunning{Bandwidth-Efficient Cooperative Perception for Heterogeneous Vehicles}

\author{Deyuan Qu\inst{1} \and Qi Chen\inst{2} \and Yongqi Zhu\inst{1} \and Yihao Zhu\inst{1} \and
Sergei S. Avedisov\inst{2} \and Song Fu\inst{1} \and Qing Yang\inst{1}}

\authorrunning{D. Qu et al.}

\institute{University of North Texas, Denton, TX, USA \and
Toyota InfoTech Labs, Mountain View, CA, USA}

\maketitle

\begin{abstract}

In cooperative perception studies,  there is often a trade-off between communication bandwidth and perception performance. While current feature fusion solutions are known for their excellent object detection performance, transmitting the entire sets of intermediate feature maps requires substantial bandwidth. Furthermore, these fusion approaches are typically limited to vehicles that use identical detection models. Our goal is to develop a solution that supports cooperative perception across vehicles equipped with different modalities of sensors. This method aims to deliver improved perception performance compared to late fusion techniques, while achieving precision similar to the state-of-art intermediate fusion, but requires an order of magnitude less bandwidth. 
  We propose HEAD, a method that fuses features from the classification and regression heads in 3D object detection networks. Our method is compatible with heterogeneous detection networks such as LiDAR PointPillars, SECOND, VoxelNet, and camera Bird's-eye View (BEV) Encoder. Given the naturally smaller feature size in the detection heads, 
  we design a self-attention mechanism to fuse the classification head and a complementary feature fusion layer to fuse the regression head. Our experiments, comprehensively evaluated on the V2V4Real and OPV2V datasets, demonstrate that HEAD is a fusion method that effectively balances communication bandwidth and perception performance.
\end{abstract}

\section{Introduction}
\label{sec:intro}
\begin{figure}[!htp]
  \begin{center}
  \includegraphics[width=4.8in, height=2in]{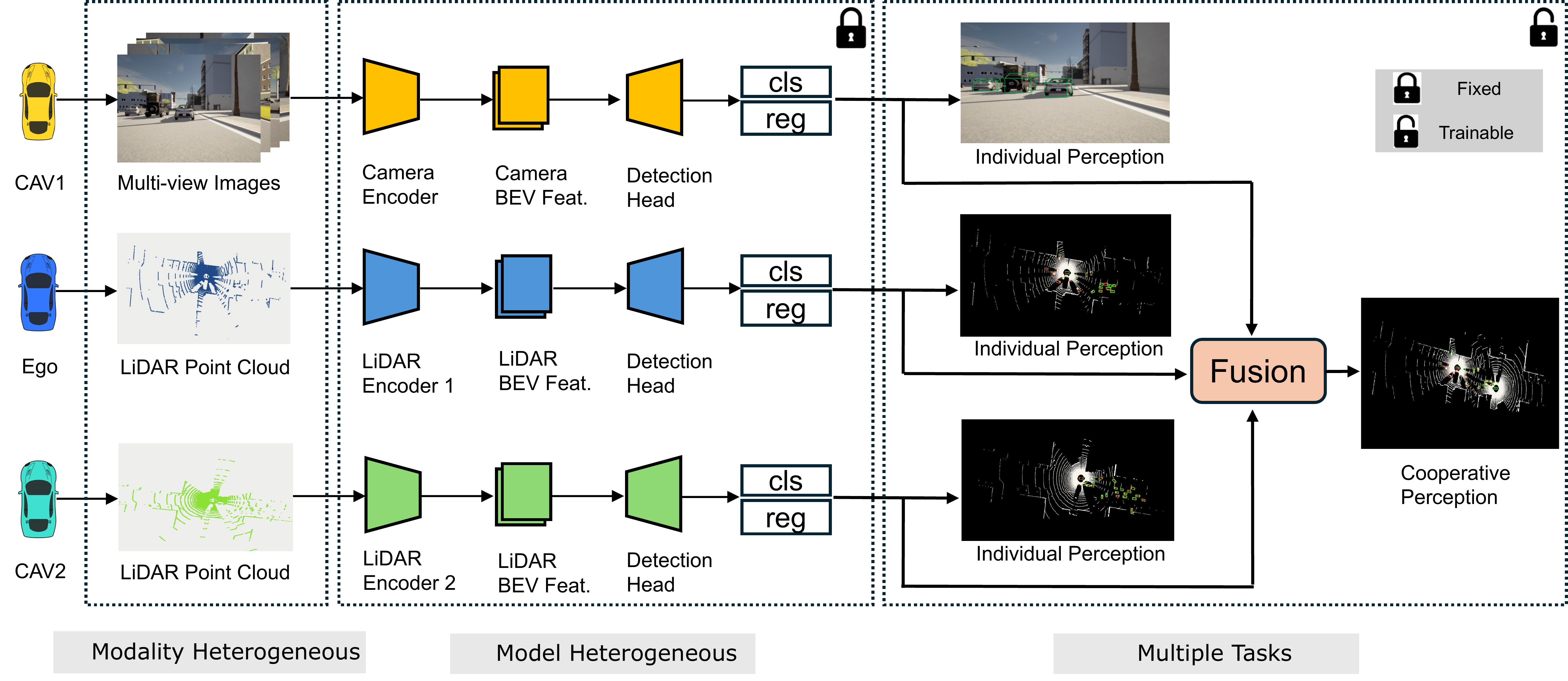}\\
   \caption{\textbf{Overview of proposed HEAD.} This is a bandwidth-efficient cooperative perception framework for heterogeneous system. All connected and autonomous vehicles can achieve effective cooperative perception without changing their individual perception performance.}
    \label{fig:idea}
    \vspace{-8mm}
  \end{center}
\end{figure}

In the fast-evolving field of cooperative perception, finding the right balance between communication bandwidth and perception performance continues to be a major challenge.
Current methods for feature fusion have gained attention due to their exceptional object detection performance compared to sharing bounding box data (Late Fusion), especially when multiple vehicles share sensing data to improve their overall perception.
However, these methods come at a high cost: the intermediate features, which encapsulate the entire sensing data, require substantial bandwidth for transmission. This presents a critical trade-off, as the increased demand for bandwidth can strain communication networks, particularly in dense traffic.

Furthermore, existing feature fusion strategies~\cite{chen2019f, xu2022opv2v, xu2022v2x, xu2022cobevt} are predominantly designed for homogeneous detection networks, where vehicles utilize identical sensing modalities and detection algorithms. 
This homogeneity limits the adaptability of cooperative perception systems in real-world scenarios, where vehicles come from different manufacturers and are often equipped with diverse sensor setups and detection networks.

Motivated by these challenges, our research aims to develop a cooperative perception method that not only enhances perception performance but also operates efficiently across vehicles with different sensor modalities. 
Our approach seeks to bridge the gap between the high bandwidth consumption of current methods and the need for robust perception across heterogeneous systems. 
Specifically, we aim to achieve a performance level that surpasses late fusion techniques while maintaining comparable bandwidth usage.

In this work, we introduce HEAD, a novel fusion method that integrates features from classification and regression heads in 3D object detection networks. 
Unlike traditional methods, HEAD is designed to be compatible with heterogeneous detection networks, including LiDAR-based systems such as PointPillars~\cite{lang2019pointpillars}, SECOND~\cite{yan2018second}, and VoxelNet~\cite{zhou2018voxelnet}, as well as camera-based Bird's-eye View (BEV) encoders. 
By focusing on the naturally smaller feature size in the detection heads, HEAD reduces the bandwidth requirements without compromising on the quality of the fused features.

To compare with existing state-of-the-art (SOTA) feature fusion methods~\cite{chen2019f, xu2022opv2v, xu2022v2x, xu2022cobevt}, we integrated a self-attention mechanism for fusing classification heads, allowing the model to dynamically balance and combine features from different sources. We also implemented a complementary feature fusion layer for regression heads, inspired by state-of-the-art techniques, to capture relevant spatial dependencies and object characteristics. Our experimental results show that our approach performs on par with state-of-the-art intermediate feature fusion methods while significantly reducing the bandwidth requirements.

The contributions of this work are as follows:
\begin{itemize}
    \item We first proposed a novel fusion method, HEAD, that integrates classification and regression heads information in 3D object detection networks.
    \item The proposed method is specifically designed to work across heterogeneous detection networks, overcoming the limitations of existing methods that require homogeneous sensor setups.
    \item The proposed method minimizes bandwidth usage, making the fusion process more efficient and better than Late Fusion.

\end{itemize}

\section{Related Work}
\subsection{Individual Perception}
LiDAR-based 3D object detection is critical for automated vehicles, with models generally split into point-based and voxel-based approaches.
Point-based methods, such as PointNet~\cite{qi2017pointnet} and PointRCNN~\cite{shi2019pointrcnn}, process raw point clouds directly, retaining fine spatial details. 
Voxel-based methods, including VoxelNet~\cite{zhou2018voxelnet}, SECOND~\cite{yan2018second} and PointPillars~\cite{lang2019pointpillars}, convert point clouds into structured grids, offering more efficient processing.
However, both approaches are limited by the constraints of vehicle sensors, which can reduce perception range and accuracy. To address these limitations, cooperative perception has emerged, where multiple vehicles share data to enhance detection performance and mitigate the challenges inherent in standalone sensor systems.

\subsection{Cooperative Perception}
Cooperative perception solutions for connected and automated vehicles can be divided into \textit{early fusion}~\cite{chen2019cooper, arnold2020cooperative}, \textit{deep fusion}~\cite{chen2019f,guo2021coff,xu2022v2x,xu2022cobevt,hu2022where2comm,lu2023robust,vnet, ma2024macp, qu2023sicp,fan2023quest,yu2022dair,dhakal2023sniffer,dhakal2024sniffer,dhakal2023virtualpainting}, and \textit{late fusion}.
Among these, the {deep fusion} achieves a good balance between bandwidth and detection performance, making it widely embraced in the research~\cite{yang2021machine}.
Some deep fusion methods have been extensively studied.
F-Cooper\cite{chen2019f} employs the \textit{maxout} operation for feature map fusion.
SiCP~\cite{qu2023sicp} proposed a simultaneous individual and cooperative perception framework, enable efficient cooperative detection while also ensuring that, in the absence of data sharing between vehicles, the detection performance remains comparable to that of standalone models.
AttFuse~\cite{xu2022opv2v} utilizes a self-attention mechanism to fuse features effectively.
V2X-ViT~\cite{xu2022v2x} introduces a unified transformer-based architecture specifically designed for multi-agent perception, while CoBEVT~\cite{xu2022cobevt} proposes a versatile transformer framework for similar applications.
For instance, Where2comm~\cite{hu2022where2comm} introduces a spatial confidence map to capture the spatial diversity of perceptual data, effectively minimizing communication bandwidth. 
Additionally, CoAlign~\cite{lu2023robust} introduces a novel hybrid collaboration framework designed to address pose errors. 
While these methods excel in cooperative perception, they primarily focus on fusing homogeneous models.
In contrast, our proposed approach not only explores heterogeneous fusion but also achieves high perception performance without requiring substantial bandwidth.

\section{Methodology}
\begin{figure}[!htp]
  \begin{center}
  \includegraphics[width=4.8in, height=1.4in]{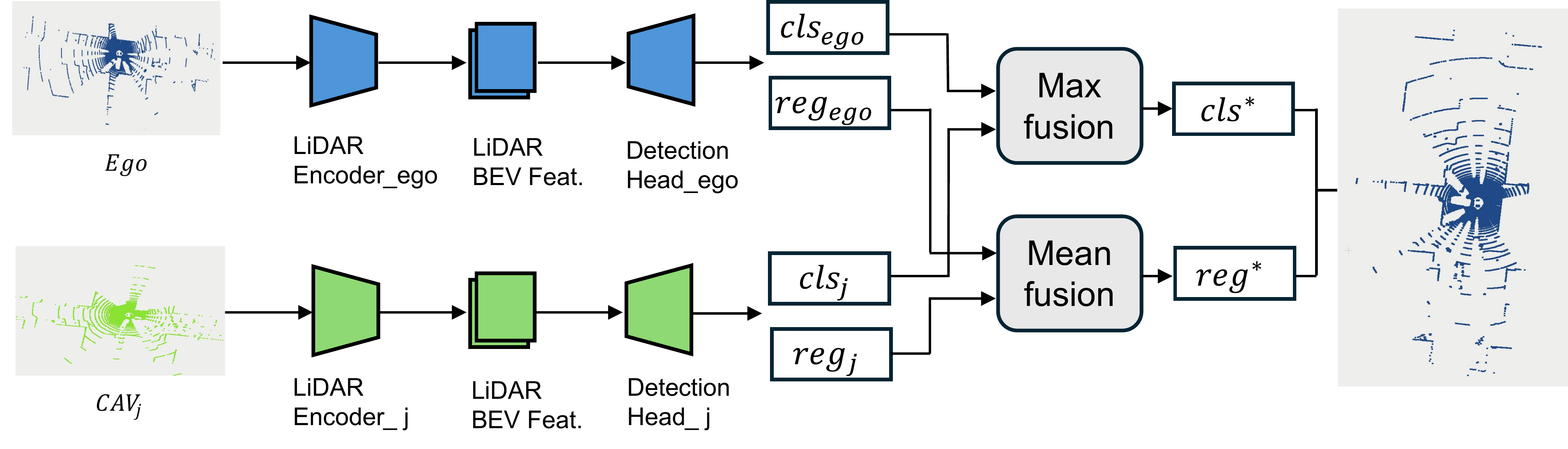}\\
   \caption{An overview of the heterogeneous fusion network. Blue Ego vehicle encoder correspond to the PointPillars backbone, whereas green $\textit{CAV}_{j}$ vehicle encoder correspond to the SECOND backbone.}
    \label{fig:heter}
    \vspace{-8mm}
  \end{center}
\end{figure}
In this section, we detail the design of our fusion network, addressing both heterogeneous and homogeneous approaches.
Section~\ref{sec:heter} focuses on the design of the heterogeneous fusion network, where we use distinct backbones for the ego vehicle and the sender vehicle. Specifically, the ego vehicle utilizes PointPillars~\cite{lang2019pointpillars}, while the sender vehicle employs SECOND~\cite{yan2018second}.
For the fusion process, we apply the Max function to merge classification maps, ensuring that the most significant features are retained. 
For regression maps, we use the Mean function to combine features, achieving a balanced integration of spatial information.
In Section~\ref{sec:homog}, we compare our method with existing homogeneous BEV feature fusion techniques. 
We employ self-attention mechanisms for fusing classification maps, effectively capturing the interdependencies between different features.
For regression map fusion, we use Complementary Feature Fusion from SiCP~\cite{qu2023sicp}, which is recognized for its robustness in handling and merging high-dimensional feature spaces. 
In this paper, we assume that feature map alignment can be effectively handled by existing solutions, such as those presented in~\cite{dong2023lidar, bbalign2024}. 
Consequently, the discussion of feature map registration is considered beyond the scope of this work. 

\begin{figure}[!htp]
  \begin{center}
  \includegraphics[width=3.8in, height=1.5in]{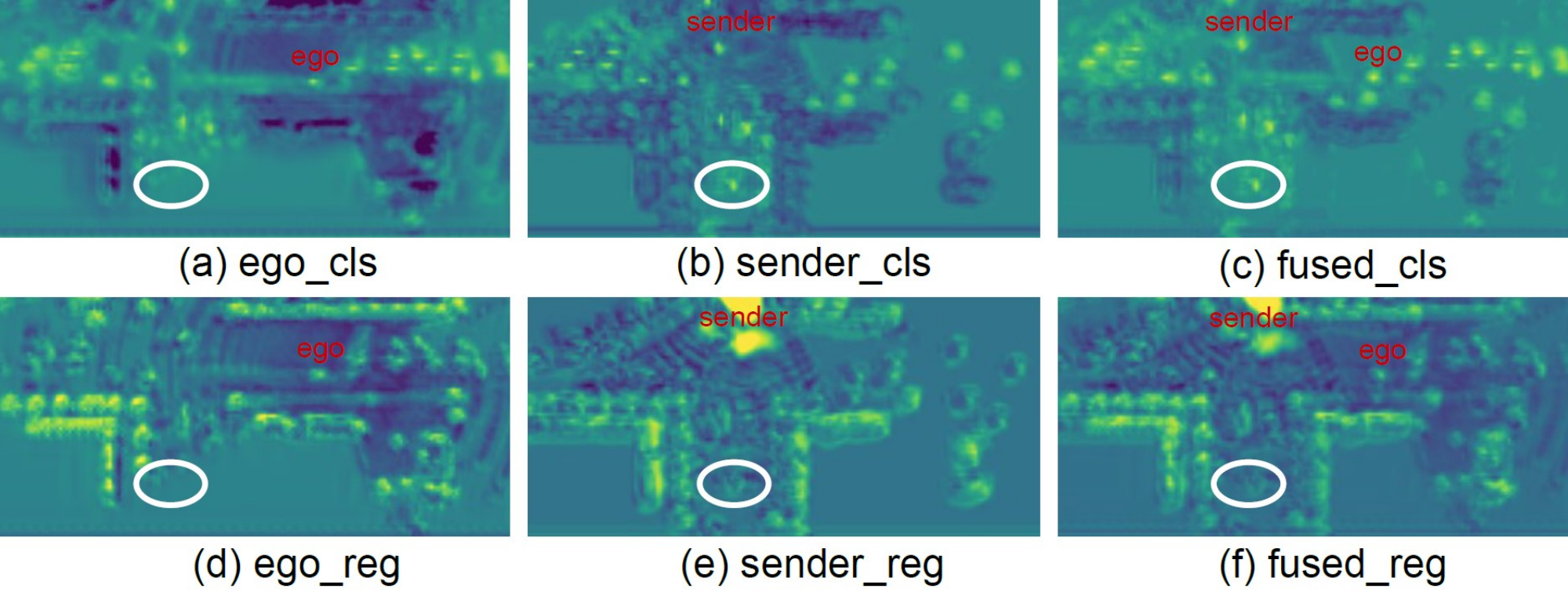}\\
   \caption{Visualizing classification and regression maps. The ego vehicle and sender vehicle use Pointpillars and SECOND to process point cloud respectively. Figure~\ref{fig:gradients}(a) and (d) represent the classification and regression maps of ego vehicle. Similarly, Figure~\ref{fig:gradients}(b) and (e) correspond to the classification and regression maps of sender vehicle.}
    \label{fig:gradients}
    \vspace{-8mm}
  \end{center}
\end{figure}

\subsection{Heterogeneous Fusion Network}
\label{sec:heter}
\noindent\textbf{Observations.} We observed several notable insights.
Figure~\ref{fig:gradients}(a), (b), and (c) show the classification maps produced by the ego vehicle using PointPillars, the sender vehicle using SECOND, and the fused classification map, respectively.
In these figures, the bright regions within the white circles indicate vehicle targets, with brighter areas suggesting a higher probability of the detected object being a vehicle.
Similarly, Figure~\ref{fig:gradients}(d), (e), and (f), show the regression maps generated by the ego vehicle using PointPillars, the sender vehicle using SECOND, and the fused regression map, respectively.
Here, the vehicle targets within the white circles exhibit a different pattern compared to the classification maps. 
Specifically, features near the vehicle's edges are more distinct, while the central regions of the vehicles show less pronounced features. 
This is due to the limited or absent LiDAR points in the empty spaces inside a car, which results in less detailed information at the center of the vehicle.

\noindent\textbf{Classification Map.} Based on our observations of the classification maps, using the max function for fusion effectively retains the most prominent target features, thereby improving the robustness of the perception system.
Let's assume the classification map of ego vehicle is $\textit{cls}_{ego}$, the classification map of vehicle $j$ is $\textit{cls}_{j}$, then the fused classification map can be shown as

\begin{equation*}
    \textit{cls}^{*} = Max (\textit{cls}_{ego}  \parallel \textit{cls}_{j}) \in \mathbb{R}^{C \times H \times W}, \tag{1}
\end{equation*}
where $\parallel$ denotes the concatenation operation.

\noindent\textbf{Regression Map.} Based on the observations of the regression maps and experimental findings, using the mean function for fusion effectively integrates information from both vehicles.
Here, the fused regression map can be expressed as

\begin{equation*}
    \textit{reg}^{*} = Mean (\textit{reg}_{ego}  \parallel \textit{reg}_{j}) \in \mathbb{R}^{C \times H \times W}, \tag{2}
\end{equation*}
where $\parallel$ denotes the concatenation operation. $\textit{reg}_{ego}$ is regression map from ego vehicle, and $\textit{reg}_{j}$ is regression map from vehicle $j$.

\subsection{Homogeneous Fusion Network}
\label{sec:homog}

\begin{figure}[!htp]
  \begin{center}
  \includegraphics[width=4.8in, height=1.4in]{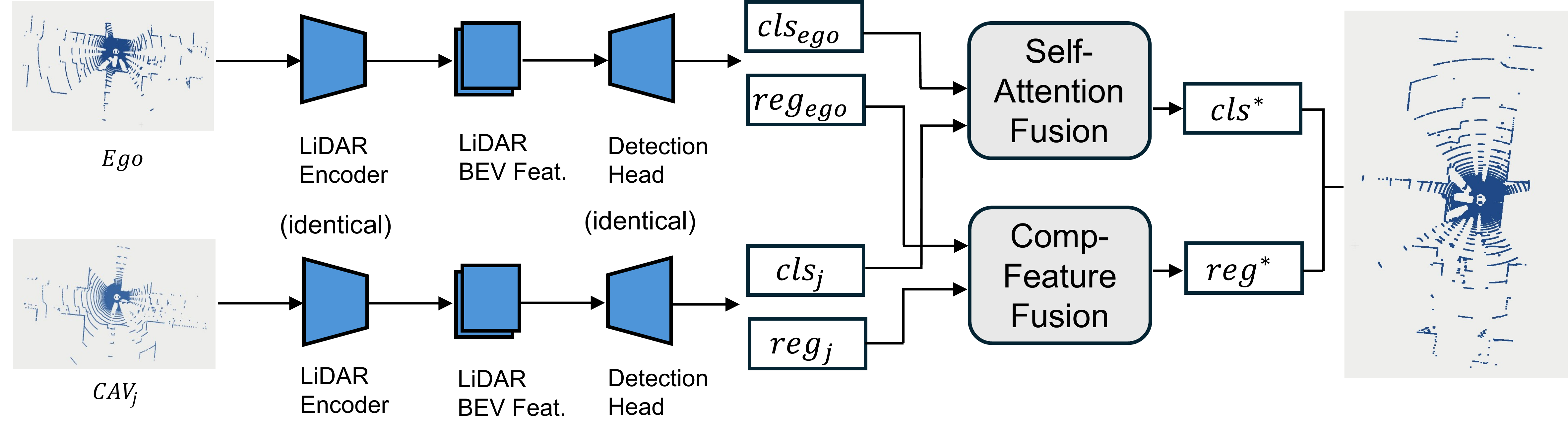}\\
   \caption{An overview of the homogeneous fusion network. Both the Ego vehicle encoder and the $\textit{CAV}_{j}$ vehicle encoder correspond to the PointPillars backbone.}
    \label{fig:homog}
    \vspace{-8mm}
  \end{center}
\end{figure}
\begin{figure*}[htbp]
    \centerline{
    \includegraphics[width=4.5in, height=1.3in]{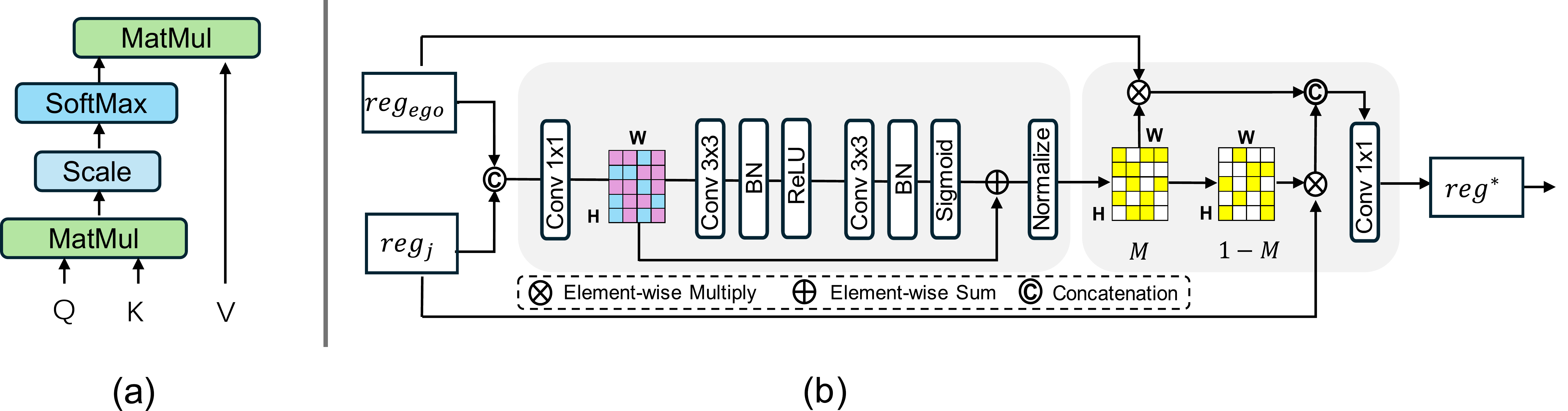}\\
    }
    \caption{Self-attention fusion and Complementary feature fusion.}
    \label{fig:cf-net}
    \vspace{-2mm}
\end{figure*}

\noindent\textbf{Classification Map.} We use a scaled dot-product attention mechanism for the fusion process~\cite{vaswani2017attention}. 
This method enables the network to assess and weigh the significance of classification maps from multiple vehicles, thereby improving overall classification accuracy.
In this approach, classification maps from various vehicles are first concatenated and then organized into matrices $Q$, $K$ and $V$. 
We compute the dot product of the query with all keys, normalize each key by $\sqrt{d_k}$, and apply the softmax function to obtain the weights for the values. 
This process is illustrated in Figure~\ref{fig:cf-net} (a).

\begin{equation}
\text{Attention}(Q, K, V) = \text{softmax}\left(\frac{QK^T}{\sqrt{d_k}}\right)V \label{eq:attention} \tag{3}
\end{equation}
The attention mechanism dynamically adjusts the contribution of each input, ensuring that the most relevant features are emphasized in the final fused output. Here we use 256 as the value of $d_k$.
%

\noindent\textbf{Regression Map.} Inspired by the SiCP~\cite{qu2023sicp}, we employ a complementary feature fusion network for the regression maps, as shown in Figure~\ref{fig:cf-net} (b).
Consider the ego vehicle receiving a regression map from a homogeneous automated vehicle $j$. 
Using the relative pose information, we transform this regression map to obtain $\text{$reg$}_{j}$.
We then concatenate the ego vehicle's regression map $\text{$reg$}_{ego}$ with the transformed map $\text{$reg$}_{j}$.
This concatenated regression map is subsequently processed through a $1 \times 1$ convolution layer.
This process can be represented as 
\begin{equation*}
    \Delta =   Conv (\text{$reg$}_{ego}  \parallel \text{$reg$}_{j}) \in \mathbb{R}^{H \times W}, \tag{4}
\end{equation*}
where $\parallel$ denotes the concatenation operation.
Here, we utilize a $1x1$ convolutional operation to handle stacked feature maps and generate a unified single-channel feature map. 
To effectively utilize the aggregated information, we perform a subsequent operation to fully capture spatial dependencies. 
Specifically, we generate a weight map $\tilde{M}\in \mathbb{R}^{H \times W}$ using the following process:
\begin{equation*}
    \tilde{M} =\Delta \oplus \left( \sigma \left( BN \left( Conv\left( \delta \left( BN \left(  Conv \left( \Delta \right) \right) \right)\right)\right)\right)\right) \tag{5}
\end{equation*}
Here \textit{BN} denotes the Batch Normalization~\cite{ioffe2015batch}, $\delta$ and $\sigma$ represent the ReLU and Sigmoid activation functions~\cite{nwankpa2018activation} respectively, and $\oplus$ indicates element-wise summation. This process involves applying two layers of $3 \times 3$ convolutional operations. 

Subsequently, we normalize $\tilde{M}$ to obtain a normalized weighted map $M$, where all values are constrained within the range $[0, 1]$.
Since $M$ is used to weight and fuse features from two separate regression maps, we adjust $M$ as follows. 
For any element $m_{ij}$ in $M$, the adjustment is given by
\begin{equation*}
     m_{ij} =
     \begin{cases}
     m_{ij}, & m_{ij} \in \text{$reg$}_{ego} \cap \text{$reg$}_{j}\\
     0, & otherwise
     \end{cases} \tag{6}
\end{equation*}
This adjustment ensures that the weight map $M$ is only applied to fuse features within the overlapping regions of $\text{$reg$}_{ego}$ and $\text{$reg$}_{j}$. For non-overlapping regions, the weight map is set to $0$, meaning that the ego vehicle relies solely on its own data for object detection in those areas.

After fusing the received regression map, the ego vehicle's regression map will be updated to $\text{$reg$}_{ego}^{*}\in \mathbb{R}^{C \times H \times W}$ is computed as follows
\begin{equation*}
 \text{$reg$}_{ego}^{*} = Conv \left(\left({M} \otimes \text{$reg$}_{ego} \right) \parallel \left( \left(1 - {M} \right) \otimes\ \text{$reg$}_{j} \right)\right)   \tag{7}
\end{equation*}
In this equation, $\otimes$ denotes element-wise multiplication. The weight map $M$ is used to adjust the ego vehicle's regression map, while the complementary weight map $(1-M)$ modifies the received regression map. As a result, in the fused regression map, each point is predominantly influenced either by the ego vehicle's information or by information from vehicle $j$.

\section{Experiments}
\label{sec:experiments}

\textbf{Datasets.}
We evaluated both heterogeneous and homogeneous cooperative perception approaches, using two widely recognized datasets: OPV2V~\cite{xu2022opv2v} and V2V4Real~\cite{xu2023v2v4real}.
OPV2V is a pioneering large-scale simulation dataset specifically designed for cooperative perception research in autonomous vehicles. It provides multi-vehicle sensor data and annotated driving scenarios, facilitating the development and testing of collaborative perception algorithms.
V2V4Real, on the other hand, is a real-world dataset collected concurrently by two vehicles, capturing a diverse range of local and highway driving scenarios.
These datasets were utilized to evaluate and benchmark Vehicle-to-Vehicle (V2V) cooperative perception methodologies.

To meet the specific training and testing requirements of our model, we implement a First-Come-First-Serve policy for each frame across all datasets in our cooperative perception tasks. This approach enables us to utilize data from both the ego vehicle and the sender vehicle.

\noindent\textbf{Implementation.}
For heterogeneous cooperative perception implementation, we adopt PointPillars~\cite{lang2019pointpillars} as ego vehicle's backbone, and SECOND~\cite{yan2018second} as sender vehicle's backbone.
We use a Element-wise Max Fusion function to fuse both classification features and use a Mean Fusion function to fuse both regression features. 
For homogeneous cooperative perception implementation, following~\cite{xu2022opv2v} setting, all the models adopt PointPillars~\cite{lang2019pointpillars} as the backbone. 
The resulting BEV feature are then passed through the detection head to produce the classification and regression features.
For multi-vehicle classification features fusion, we utilize a self-attention mechanism~\cite{xu2022opv2v} to enhance collaborative perception. The multi-vehicle regression features are fused using a complementary feature fusion approach~\cite{qu2023sicp}. Our model was trained using an Nvidia RTX 3090 GPU, with the Adam optimizer, a learning rate of 0.001, and a batch size of 1.
At the inference stage, following~\cite{xu2022opv2v,xu2023v2v4real}, we use the Average Precision (AP) metric to evaluate the performance of all models on OPV2V and V2V4real datasets. 
The evaluation involved employing IoU thresholds of 50 and 70, respectively.

\noindent\textbf{Baselines.}
To compare against our proposed HEAD method, we adopted the following baseline models:
For heterogeneous cooperative perception, we compare our model with no fusion and late fusion.
For homogeneous cooperative perception, several SOTA deep fusion models (including F-Cooper~\cite{chen2019f}, AttFuse~\cite{xu2022opv2v}, V2X-ViT~\cite{xu2022v2x} and CoBEVT~\cite{xu2022cobevt}) and late-fusion.

\begin{figure}[!htp]
  \centering
  \begin{subfigure}{.5\columnwidth}
    \centering
    \includegraphics[height=1in, width=1.8in]{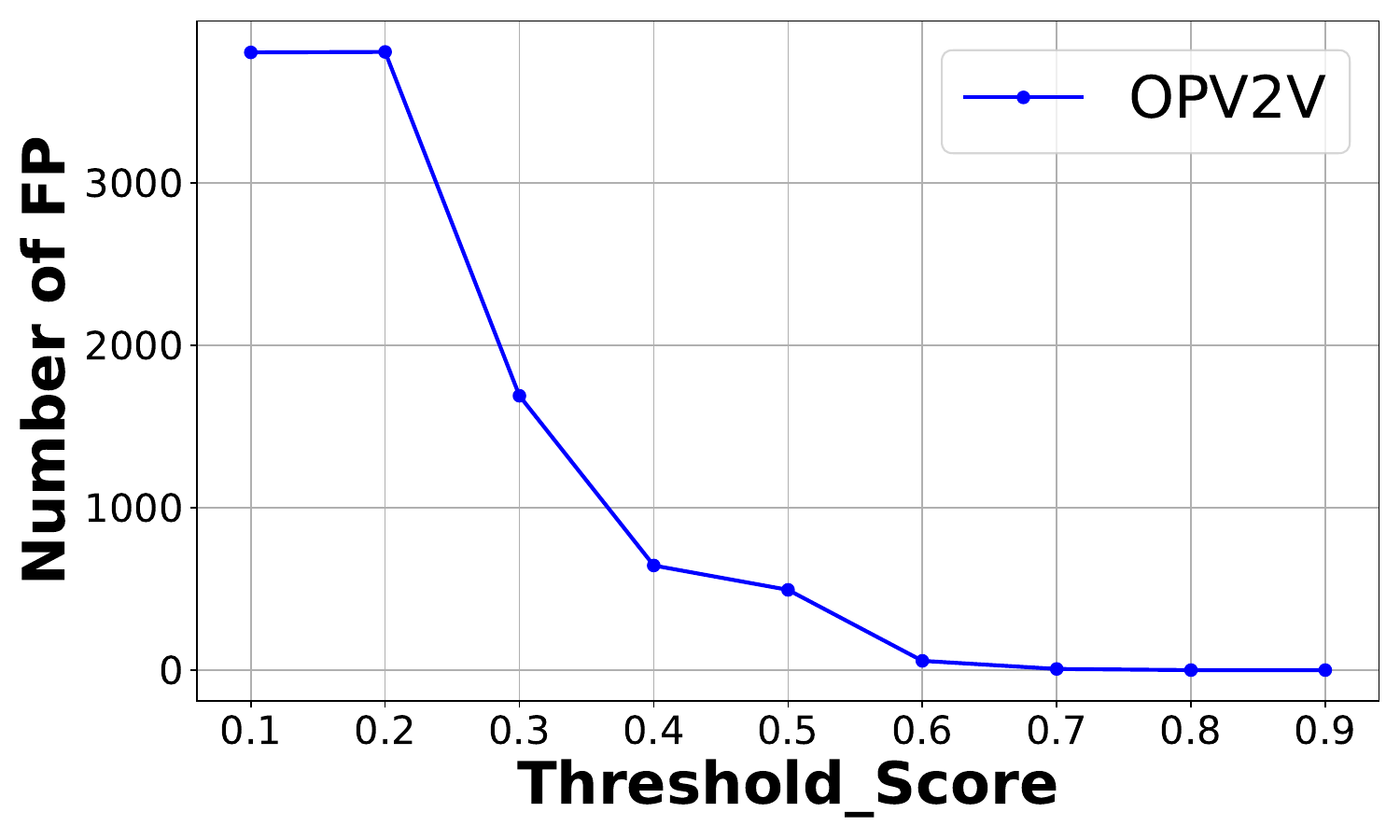} 
    \caption{On OPV2V Dataset}
    \label{fig:sub2}
  \end{subfigure}%
  \begin{subfigure}{.5\columnwidth} 
    \centering
    \includegraphics[height=1in, width=1.8in]{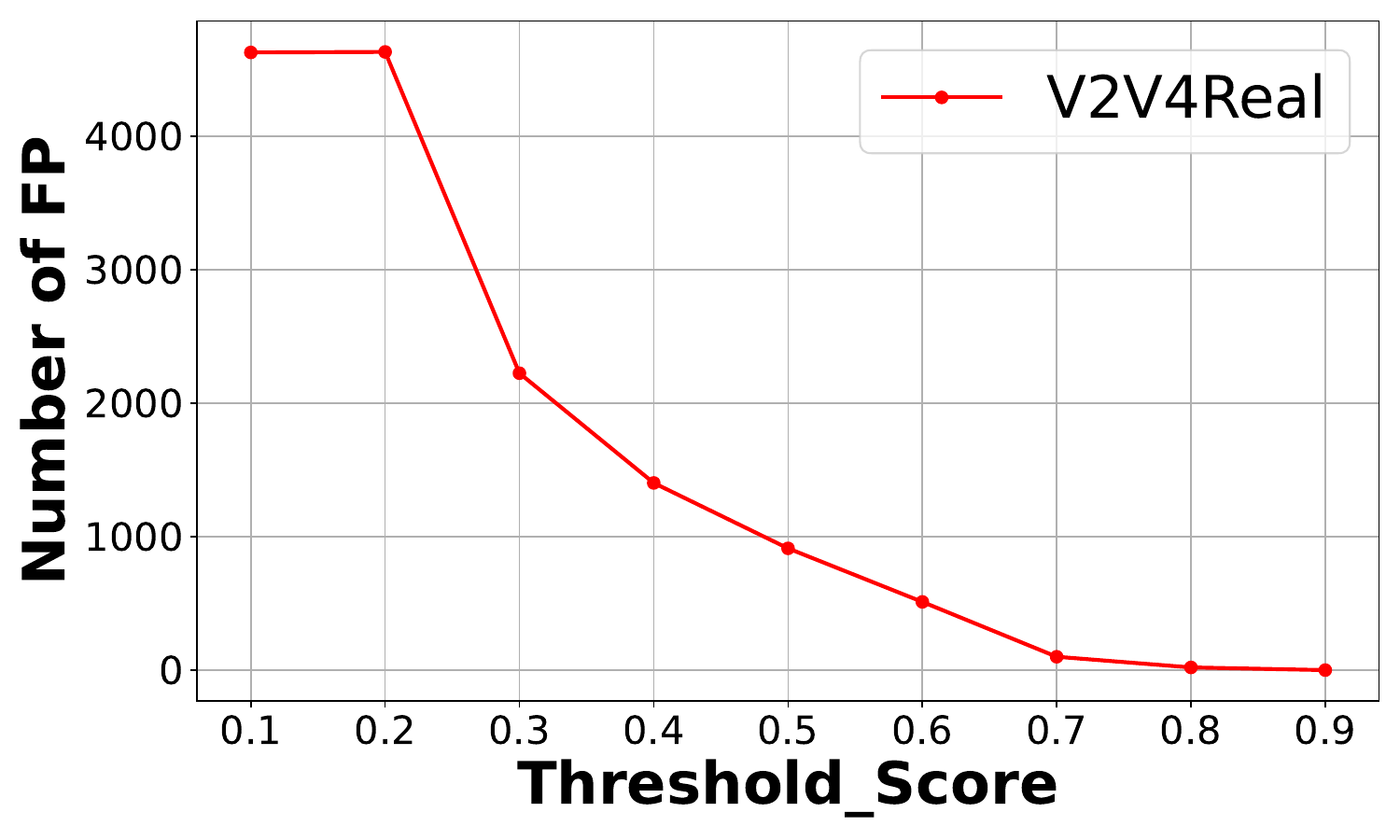} 
    \caption{On V2V4Real Dataset}
    \label{fig:sub1}
  \end{subfigure}
  \caption{Impact of threshold score on false positives across datasets for late fusion.}
  \label{fig:fp vs threshold score}
  \vspace{-8mm}
\end{figure}

\noindent\textbf{Late Fusion Evaluation Criteria.} Particular attention is given to late-fusion, which aggregates the final predictions from multiple vehicles at a later stage, we require that the detection results transmitted by the sender vehicle must have high confidence.
This means that the transmission objects should have higher scores so that all sent objects are true positives (TP).
Experiments on both datasets show, as depicted in Figure~\ref{fig:fp vs threshold score}, that when the detection score threshold is set to 0.75, there are no false positives (FP) in the transmitted objects, ensuring reliability.
Of course, the threshold can be adjusted according to specific application requirements.

\begin{table}[tb]
  \caption{Heterogeneous Cooperative Perception Average Precision (AP) Evaluation on OPV2V
Datasets. }
  \label{tab:heterogeneous}
  \centering
  \setlength{\tabcolsep}{12pt} 
  \begin{tabular}{@{}>{\centering\arraybackslash}m{3cm}|c|c|c|c@{}}
    \toprule
    \multirow{2}{*}{\textbf{Method}} & \multicolumn{2}{c|}{\textbf{Default Towns}} & \multicolumn{2}{c}{\textbf{Culver City}}\\
    \cmidrule(lr){2-5}
    & AP50$ \uparrow $ & AP70$ \uparrow $ & AP50$ \uparrow $ & AP70$ \uparrow $ \\
    \midrule
    No Fusion  & 73.84 & 60.89 & 77.71 & 62.21\\
    Late Fusion & 77.31 & 63.51 & 75.62 & 58.54\\
    \midrule
    \textbf{HEAD (Ours)} & 81.11 & 64.41 & 83.39 & 64.56\\
    \bottomrule
  \end{tabular}
\end{table}

\begin{table}[htbp]
\centering
\caption{Homogeneous Cooperative Perception Evaluation on V2V4Real and OPV2V Datasets. BW means Bandwidth.}
\label{tab:homogeneous}
\footnotesize %
\setlength{\tabcolsep}{3pt} 
\resizebox{\linewidth}{!}{ 
\begin{tabular}{
  >{\centering\arraybackslash}m{2.5cm}|
  >{\centering\arraybackslash}m{1.5cm}|
  >{\centering\arraybackslash}m{2.2cm}|
  >{\centering\arraybackslash}m{2cm}|
  >{\centering\arraybackslash}m{1.5cm}|
  >{\centering\arraybackslash}m{2.2cm}|
  >{\centering\arraybackslash}m{2cm}|
  >{\centering\arraybackslash}m{1.5cm}}
\toprule
\rowcolor{lightgray}
 &  & \multicolumn{3}{c|}{\textbf{V2V4Real Dataset}} & \multicolumn{3}{c}{\textbf{OPV2V Default Dataset}} \\ 
\cline{3-8}\arrayrulecolor{black}
\rowcolor{lightgray}
 \textbf{Method}& \textbf{Channels} & \textbf{BW(Mbps) $ \downarrow $} & \textbf{Ave Infer Time(ms) $ \downarrow $} & \textbf{AP50 $ \uparrow $} & \textbf{BW(Mbps) $ \downarrow $} & \textbf{Ave Infer Time(ms) $ \downarrow $} & \textbf{AP50 $ \uparrow $} \\
\midrule
F-Cooper~\cite{chen2019f} & 256 & 660 & 15.16 & 47.7 & 2749.6 & 27.48 & 79.71 \\
Attfuse~\cite{xu2022opv2v} & 256 & 660 & 15.24 & 43.0 & 2749.6 & 31.46 & 81.91 \\
V2X-ViT~\cite{xu2022v2x} & 256 & 660 & 43.51 & 49.2 & 2749.6 & 97.54 & 85.62 \\
CoBEVT~\cite{xu2022cobevt} & 256 & 660 & 27.53 & 51.0 & 2749.6 & 58.13 & 84.77 \\
\midrule
Late Fusion & - & 0.09 & 15.25 & 40.4 & 0.37 & 31.45 & 77.03 \\
\midrule
\textbf{HEAD (Ours)} & 16 & 41.6 & 17.67 & 48.4 & 172 & 33.50 & 81.88 \\
\bottomrule
\end{tabular}
}
\end{table}
\vspace{-10pt}
\subsection{Heterogeneous Cooperative Perception Evaluations}
For No Fusion (the ego vehicle relies solely on it's LiDAR measurements without cooperation), ego vehicle use PointPillars as backbone. For Late Fusion and HEAD(Ours), ego vehicle use PointPillars as backbone, sender vehicle use SECOND as backbone.
As depicted in Table~\ref{tab:heterogeneous} shows that the proposed HEAD method significantly outperforms both No Fusion and Late Fusion approaches in heterogeneous cooperative perception on the OPV2V datasets. 
For example, in Default Towns, HEAD achieves an AP50 of 81.11 and an AP70 of 64.41, compared to 73.84 and 60.89 for No Fusion, and 77.31 and 63.51 for Late Fusion. Similarly, in Culver City, HEAD reaches an AP50 of 83.39 and an AP70 of 64.56, surpassing No Fusion (77.71, 62.21) and Late Fusion (75.62, 58.54). 
These results indicate that HEAD not only improves detection accuracy by up to 7.27\% for AP50 and 5.87\% for AP70 compared to the baseline but also delivers more consistent performance across different environments.

\subsection{Homogeneous Cooperative Perception Evaluations}
\noindent\textbf{Evaluation on OPV2V dataset.} In the Table~\ref{tab:homogeneous}, HEAD (Ours) significantly reduces bandwidth consumption to 172 Mbps, a substantial improvement compared to other intermediate fusion methods, which exceed 2749 Mbps. This demonstrates its superior efficiency in collaborative sensing.
HEAD also achieves an average inference time of 33.5 ms, outperforming all methods except for the Late Fusion model.
In terms of performance, HEAD attains an AP50 (Average Precision at 50\% IoU) score of 81.88, surpassing the Late Fusion strategy and approaching the level of intermediate fusion methods.

\noindent\textbf{Evaluation on V2V4Real dataset.} Similar to the OPV2V dataset, HEAD (Ours) model significantly reduces the bandwidth required for data transmission to 41.6 Mbps compared to other intermediate fusion methods that require around 660 Mbps. This reduction shows a substantial improvement in communication efficiency.
With an average inference time of 17.67 ms, HEAD (Ours) maintains competitive processing speeds, being only slightly slower than some intermediate methods (like F-Cooper, AttFuse, and Late Fusion) but much faster than V2X-ViT and CoBEVT.
HEAD achieves an AP50 (Average Precision at 50\% IoU) score of 48.4, which is higher than the Late Fusion approach and close to the scores of other intermediate fusion methods.
\begin{figure}[!htp]
  \begin{center}
  \includegraphics[width=3.5in, height=1.3in]{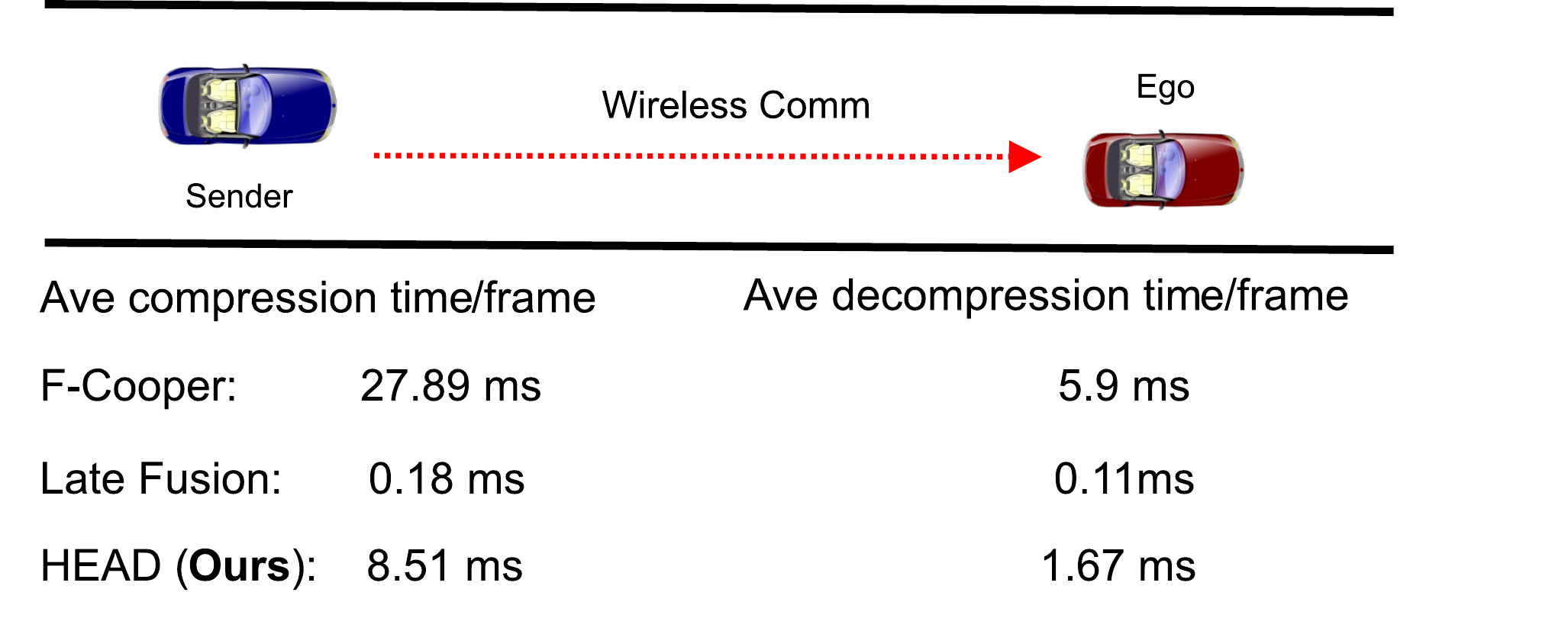}\\
   \caption{Comparison of information average compression times and average decompression times per frame. Including F-Cooper, Late Fusion, and HEAD (Ours).}
    \label{fig:data transmission}
    \vspace{-8mm}
  \end{center}
\end{figure}

\noindent\textbf{Data Transmission Analysis.} To provide a comprehensive analysis of bandwidth and transmission efficiency, we evaluated three fusion strategies: F-Cooper, Late Fusion, and HEAD.
As shown in Figure~\ref{fig:data transmission}, we further reduced bandwidth requirements by compressing feature maps and detection results before transmission from the sender to the ego vehicle. 
The ego vehicle then decompresses the received information.
The experimental results reveal that Late Fusion requires less than 1 ms in total due to its minimal data volume. 
In contrast, F-Cooper, representing state-of-the-art models, takes 33.79 ms, while HEAD achieves a faster processing time of only 10.18 ms. 
Considering the high demands for real-time performance and detection accuracy in connected and autonomous vehicles, and excluding the wireless communication time, HEAD demonstrates exceptional bandwidth efficiency, competitive inference times, and strong detection accuracy across both datasets, highlighting its significant importance.

\subsection{Qualitative Evaluations}
\begin{figure*}[!htp]
  \begin{center}
  \includegraphics[width=4.8in, height=1in]{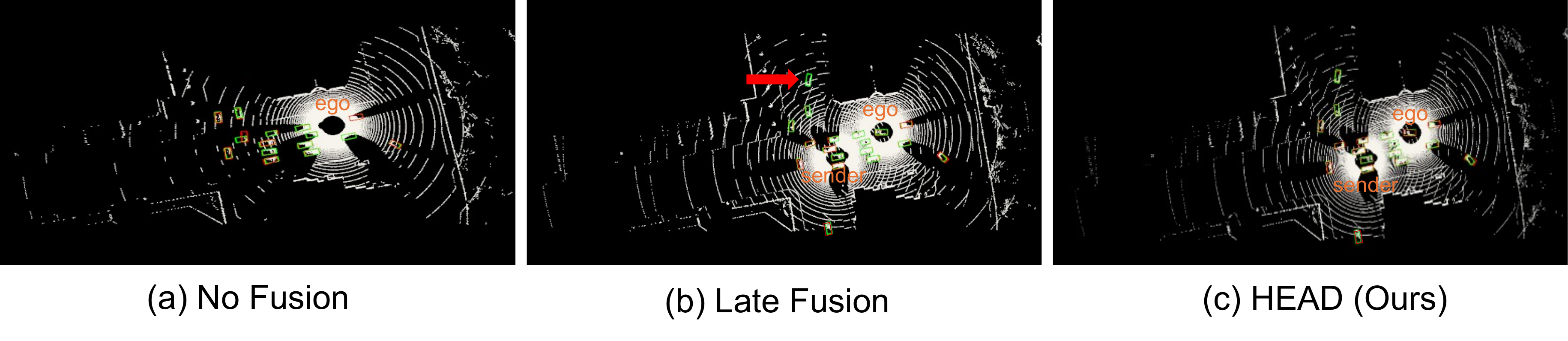}\\
   \caption{Qualitative evaluations on heterogeneous cooperative perception.}
    \label{fig:qual-heter}
    \vspace{-13mm}
  \end{center}
\end{figure*}

As shown in Figure~\ref{fig:qual-heter}, the ego vehicle utilizes PointPillars as its backbone, while the sender vehicle employs SECOND as its backbone. The object pointed by the \textcolor{red}{red arrow} is not detected in Figure~\ref{fig:qual-heter} (b), whereas the same object in the same position in Figure~\ref{fig:qual-heter} (c) is successfully detected. The reason is that in Late fusion, the detection score of the object by the sender vehicle not high enough, resulting in it not being transmitted to the ego vehicle. However, after fusion through the HEAD method, the object is detected. This highlights the importance of multi-vehicle collaboration and fusion strategies in improving detection performance, especially for low-confidence objects.

\begin{figure*}[!htp]
  \begin{center}
  \includegraphics[width=4.8in, height=3in]{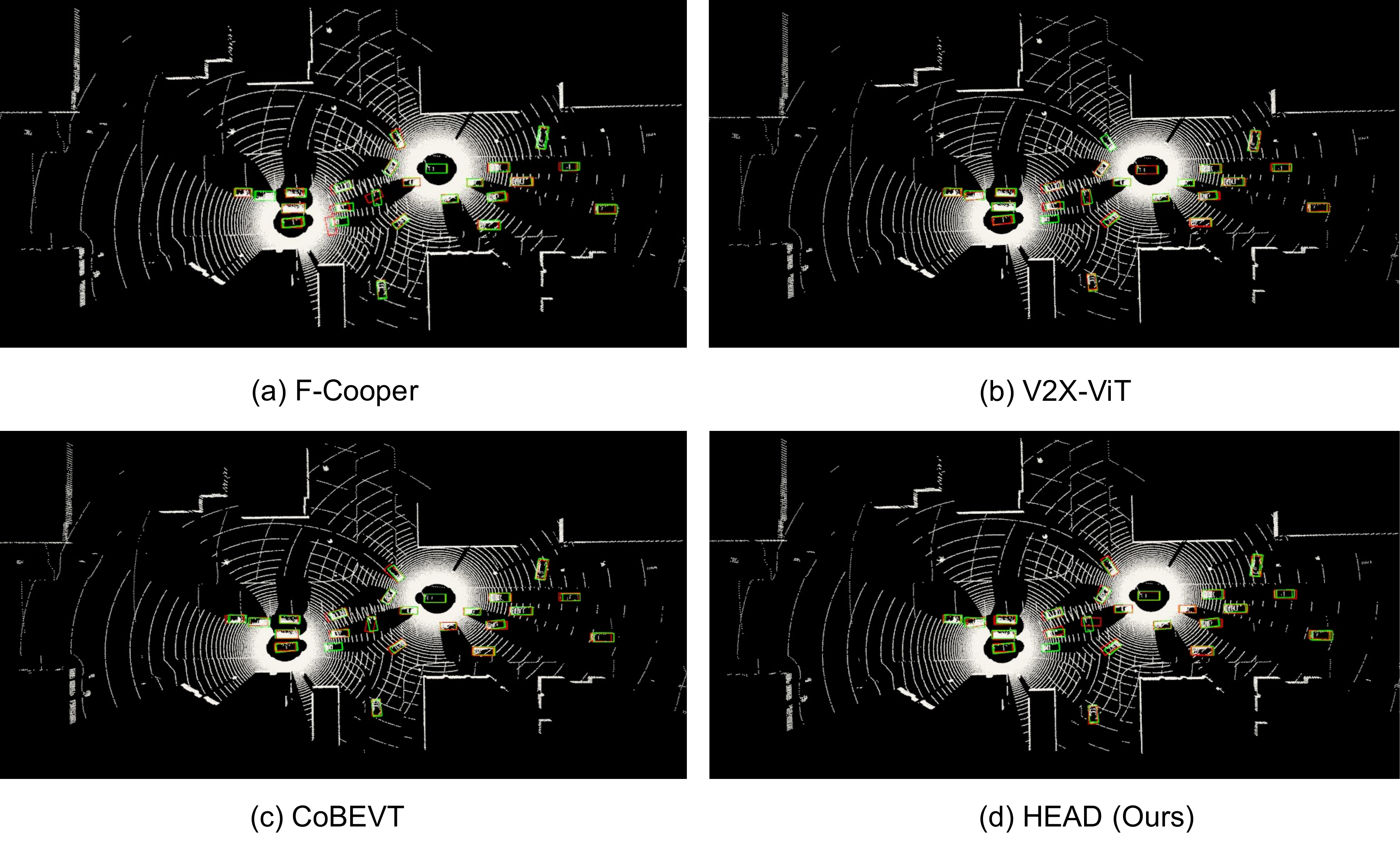}\\
   \caption{Qualitative evaluations on homogeneous cooperative perception.}
    \label{fig:qual-homo}
  \end{center}
  \vspace{-8mm}
\end{figure*}

Figure~\ref{fig:qual-homo} shows a comparison between HEAD and existing SOTA methods using homologous models, where all models utilize PointPillars as the backbone. It can be observed that our method outperforms F-Cooper and is slightly less effective than V2X-ViT and CoBEVT. However, it is important to note that our method requires significantly less bandwidth compared to these other approaches. This trade-off highlights the efficiency of our method in scenarios where bandwidth constraints are a critical factor, while still maintaining competitive performance levels in detection accuracy.
%
\section{Conclusion}
In this paper, we present HEAD, a fusion method designed for both heterogeneous and homogeneous detection networks. 
Unlike traditional methods, HEAD integrates features from the classification and regression heads of various heterogeneous detection frameworks, such as LiDAR and camera-based systems (with camera-based experiments to be added in future work), efficiently and without excessive bandwidth consumption.
Furthermore, by employing self-attention mechanisms and complementary fusion layers in homogeneous networks, HEAD achieves an effective balance between detection accuracy and communication efficiency. 
Evaluations on the V2V4Real and OPV2V datasets demonstrate that HEAD not only delivers high performance but also significantly reduces bandwidth usage, making it a practical solution for cooperative perception in diverse environments.

%
%
\bibliographystyle{splncs04}
\bibliography{main}

\begin{thebibliography}{10}
\providecommand{\url}[1]{\texttt{#1}}
\providecommand{\urlprefix}{URL }
\providecommand{\doi}[1]{https://doi.org/#1}

\bibitem{arnold2020cooperative}
Arnold, E., Dianati, M., de~Temple, R., Fallah, S.: Cooperative perception for 3d object detection in driving scenarios using infrastructure sensors. IEEE Transactions on Intelligent Transportation Systems  \textbf{23}(3),  1852--1864 (2020)

\bibitem{chen2019f}
Chen, Q., Ma, X., Tang, S., Guo, J., Yang, Q., Fu, S.: F-cooper: Feature based cooperative perception for autonomous vehicle edge computing system using 3d point clouds. In: Proceedings of the 4th ACM/IEEE Symposium on Edge Computing. pp. 88--100 (2019)

\bibitem{chen2019cooper}
Chen, Q., Tang, S., Yang, Q., Fu, S.: Cooper: Cooperative perception for connected autonomous vehicles based on 3d point clouds. In: 2019 IEEE 39th International Conference on Distributed Computing Systems (ICDCS). pp. 514--524. IEEE (2019)

\bibitem{dhakal2023virtualpainting}
Dhakal, S., Carrillo, D., Qu, D., Nutt, M., Yang, Q., Fu, S.: Virtualpainting: Addressing sparsity with virtual points and distance-aware data augmentation for 3d object detection. arXiv preprint arXiv:2312.16141  (2023)

\bibitem{dhakal2024sniffer}
Dhakal, S., Carrillo, D., Qu, D., Yang, Q., Fu, S.: Sniffer faster r-cnn++: An efficient camera-lidar object detector with proposal refinement on fused candidates. Journal on Autonomous Transportation Systems  \textbf{1}(2),  1--18 (2024)

\bibitem{dhakal2023sniffer}
Dhakal, S., Chen, Q., Qu, D., Carillo, D., Yang, Q., Fu, S.: Sniffer faster r-cnn: A joint camera-lidar object detection framework with proposal refinement. In: 2023 IEEE International Conference on Mobility, Operations, Services and Technologies (MOST). pp. 1--10. IEEE (2023)

\bibitem{dong2023lidar}
Dong, J., Chen, Q., Qu, D., Lu, H., Ganlath, A., Yang, Q., Chen, S., Labi, S.: Lidar-based cooperative relative localization. In: 2023 IEEE Intelligent Vehicles Symposium (IV). pp.~1--8. IEEE (2023)

\bibitem{fan2023quest}
Fan, S., Yu, H., Yang, W., Yuan, J., Nie, Z.: Quest: Query stream for vehicle-infrastructure cooperative perception. arXiv preprint arXiv:2308.01804  (2023)

\bibitem{guo2021coff}
Guo, J., Carrillo, D., Tang, S., Chen, Q., Yang, Q., Fu, S., Wang, X., Wang, N., Palacharla, P.: Coff: Cooperative spatial feature fusion for 3-d object detection on autonomous vehicles. IEEE Internet of Things Journal  \textbf{8}(14),  11078--11087 (2021)

\bibitem{hu2022where2comm}
Hu, Y., Fang, S., Lei, Z., Zhong, Y., Chen, S.: Where2comm: Communication-efficient collaborative perception via spatial confidence maps. Advances in neural information processing systems  \textbf{35},  4874--4886 (2022)

\bibitem{ioffe2015batch}
Ioffe, S., Szegedy, C.: Batch normalization: Accelerating deep network training by reducing internal covariate shift. In: International conference on machine learning. pp. 448--456. pmlr (2015)

\bibitem{lang2019pointpillars}
Lang, A.H., Vora, S., Caesar, H., Zhou, L., Yang, J., Beijbom, O.: Pointpillars: Fast encoders for object detection from point clouds. In: Proceedings of the IEEE/CVF conference on computer vision and pattern recognition. pp. 12697--12705 (2019)

\bibitem{lu2023robust}
Lu, Y., Li, Q., Liu, B., Dianati, M., Feng, C., Chen, S., Wang, Y.: Robust collaborative 3d object detection in presence of pose errors. In: 2023 IEEE International Conference on Robotics and Automation (ICRA). pp. 4812--4818. IEEE (2023)

\bibitem{ma2024macp}
Ma, Y., Lu, J., Cui, C., Zhao, S., Cao, X., Ye, W., Wang, Z.: Macp: Efficient model adaptation for cooperative perception. In: Proceedings of the IEEE/CVF Winter Conference on Applications of Computer Vision. pp. 3373--3382 (2024)

\bibitem{nwankpa2018activation}
Nwankpa, C., Ijomah, W., Gachagan, A., Marshall, S.: Activation functions: Comparison of trends in practice and research for deep learning. arXiv preprint arXiv:1811.03378  (2018)

\bibitem{qi2017pointnet}
Qi, C.R., Su, H., Mo, K., Guibas, L.J.: Pointnet: Deep learning on point sets for 3d classification and segmentation. In: Proceedings of the IEEE conference on computer vision and pattern recognition. pp. 652--660 (2017)

\bibitem{qu2023sicp}
Qu, D., Chen, Q., Bai, T., Qin, A., Lu, H., Fan, H., Fu, S., Yang, Q.: Sicp: Simultaneous individual and cooperative perception for 3d object detection in connected and automated vehicles. arXiv preprint arXiv:2312.04822  (2023)

\bibitem{shi2019pointrcnn}
Shi, S., Wang, X., Li, H.: Pointrcnn: 3d object proposal generation and detection from point cloud. In: Proceedings of the IEEE/CVF conference on computer vision and pattern recognition. pp. 770--779 (2019)

\bibitem{bbalign2024}
Song, L., Valentine, W., Yang, Q., Wang, H., Fang, H., Liu, Y.: Bb-align: A lightweight pose recovery framework for vehicle-to-vehicle cooperative perception. In: The 44th IEEE International Conference on Distributed Computing Systems (ICDCS). IEEE (2024)

\bibitem{vaswani2017attention}
Vaswani, A.: Attention is all you need. arXiv preprint arXiv:1706.03762  (2017)

\bibitem{vnet}
Wang, T.H., Manivasagam, S., Liang, M., Yang, B., Zeng, W., Urtasun, R.: V2vnet: Vehicle-to-vehicle communication for joint perception and prediction. In: European Conference on Computer Vision. pp. 605--621. Springer (2020)

\bibitem{xu2022cobevt}
Xu, R., Tu, Z., Xiang, H., Shao, W., Zhou, B., Ma, J.: Cobevt: Cooperative bird's eye view semantic segmentation with sparse transformers. arXiv preprint arXiv:2207.02202  (2022)

\bibitem{xu2023v2v4real}
Xu, R., Xia, X., Li, J., Li, H., Zhang, S., Tu, Z., Meng, Z., Xiang, H., Dong, X., Song, R., et~al.: V2v4real: A real-world large-scale dataset for vehicle-to-vehicle cooperative perception. In: Proceedings of the IEEE/CVF Conference on Computer Vision and Pattern Recognition. pp. 13712--13722 (2023)

\bibitem{xu2022v2x}
Xu, R., Xiang, H., Tu, Z., Xia, X., Yang, M.H., Ma, J.: V2x-vit: Vehicle-to-everything cooperative perception with vision transformer. In: European conference on computer vision. pp. 107--124. Springer (2022)

\bibitem{xu2022opv2v}
Xu, R., Xiang, H., Xia, X., Han, X., Li, J., Ma, J.: Opv2v: An open benchmark dataset and fusion pipeline for perception with vehicle-to-vehicle communication. In: 2022 International Conference on Robotics and Automation (ICRA). pp. 2583--2589. IEEE (2022)

\bibitem{yan2018second}
Yan, Y., Mao, Y., Li, B.: Second: Sparsely embedded convolutional detection. Sensors  \textbf{18}(10), ~3337 (2018)

\bibitem{yang2021machine}
Yang, Q., Fu, S., Wang, H., Fang, H.: Machine-learning-enabled cooperative perception for connected autonomous vehicles: Challenges and opportunities. IEEE Network  \textbf{35}(3),  96--101 (2021)

\bibitem{yu2022dair}
Yu, H., Luo, Y., Shu, M., Huo, Y., Yang, Z., Shi, Y., Guo, Z., Li, H., Hu, X., Yuan, J., et~al.: Dair-v2x: A large-scale dataset for vehicle-infrastructure cooperative 3d object detection. In: Proceedings of the IEEE/CVF Conference on Computer Vision and Pattern Recognition. pp. 21361--21370 (2022)

\bibitem{zhou2018voxelnet}
Zhou, Y., Tuzel, O.: Voxelnet: End-to-end learning for point cloud based 3d object detection. In: Proceedings of the IEEE conference on computer vision and pattern recognition. pp. 4490--4499 (2018)

\end{thebibliography}
\end{document}